\documentclass[10pt,twocolumn,letterpaper]{article}

\usepackage[pagenumbers]{cvpr} 

\usepackage{graphicx}
\usepackage{amsmath}
\usepackage{amssymb}
\usepackage{booktabs}
\usepackage{xcolor}
\usepackage{tabularx}

\usepackage[pagebackref,breaklinks,colorlinks]{hyperref}

\usepackage[capitalize]{cleveref}
\crefname{section}{Sec.}{Secs.}
\Crefname{section}{Section}{Sections}
\Crefname{table}{Table}{Tables}
\crefname{table}{Tab.}{Tabs.}

\def\cvprPaperID{*****} 
\def\confName{CVPR}
\def\confYear{2022}

\begin{document}

\title{Tell Me the Evidence? Dual Visual-Linguistic Interaction for Answer Grounding}

\author{
Junwen Pan\textsuperscript{\rm 1},
Guanlin Chen\textsuperscript{\rm 2}, Yi Liu\textsuperscript{\rm 2},
Jiexiang Wang\textsuperscript{\rm 1}, 
Cheng Bian\textsuperscript{\rm 1}, 
Pengfei Zhu\textsuperscript{\rm 2}, Zhicheng Zhang\textsuperscript{\rm 1}
\\
\textsuperscript{\rm 1}Xiaohe Healthcare, ByteDance, Guangzhou, China\\
\textsuperscript{\rm 2}College of Intelligence and Computing, Tianjin University, Tianjin, China\\
{\tt\small panjunwen@bytedance.com}
}
\maketitle


\begin{abstract}
   Answer grounding aims to reveal the visual evidence for visual question answering (VQA), which entails highlighting relevant positions in the image when answering questions about images. 
   Previous attempts typically tackle this problem using pretrained object detectors, but without the flexibility for objects not in the predefined vocabulary. 
   Recent visual-linguistic models have made remarkable advances by leveraging powerful Transformers to enable visual-linguistic interaction.
   However, these black-box methods solely concentrate on the linguistic generation, ignoring the visual interpretability.
   In this paper, we propose \textbf{D}u\textbf{a}l \textbf{V}isual-Linguistic \textbf{I}nteraction (\textbf{DaVI}), a novel unified end-to-end framework with the capability for both linguistic answering and visual grounding. 
   DaVI innovatively introduces two visual-linguistic interaction mechanisms: 1) visual-based linguistic encoder that understands questions incorporated with visual features and produces linguistic-oriented evidence for further answer decoding, and 2) linguistic-based visual decoder that focuses visual features on the evidence-related regions for answer grounding.
   This way, our approach ranked the 1$^{st}$ place in the answer grounding track of 2022 VizWiz Grand Challenge.
\end{abstract}

\section{Introduction}

Visual question answering (VQA) is a fundamental visual-linguistic task in various real-life applications such as assisting visually impaired people to answer questions about images~\cite{chen2022grounding}.
Although the VQA community has made significant progress, the best-performing systems are complicated black-box models, raising concerns about whether their answer reasoning is based on correct visual evidence.
By understanding the reasoning mechanism of the model, we can evaluate the quality of answers, improve model performance, and provide explanations for end-users.


To address this problem, answer grounding has been introduced into VQA systems, which requires the model to locate relevant image regions as well as answer visual questions.
Most previous efforts usually rely on pretrained detection models to provide object features, entailing low answering flexibility for objects not in the predefined vocabulary~\cite{KhanKDGLS21:found}. 
Recently, a few works try to get rid of these limitations in a weakly supervised fashion~\cite{KhanKDGLS21:found}, but neither the answering accuracy nor grounding accuracy achieved is satisfactory~\cite{chen2022grounding}.
In contrast, visual language pre-training has tremendously advanced VQA performance using large-scale multi-modal Transformers and web-scale datasets~\cite{li2022blip}, eliminating the need for detectors.
However, these text-generation oriented schemes concentrate on the linguistic modeling based on off-the-shelf visual features, and are therefore not capable of visual-oriented prediction in the grounding task.

\begin{figure}[!t]
  \centering
   \includegraphics[width=1\linewidth]{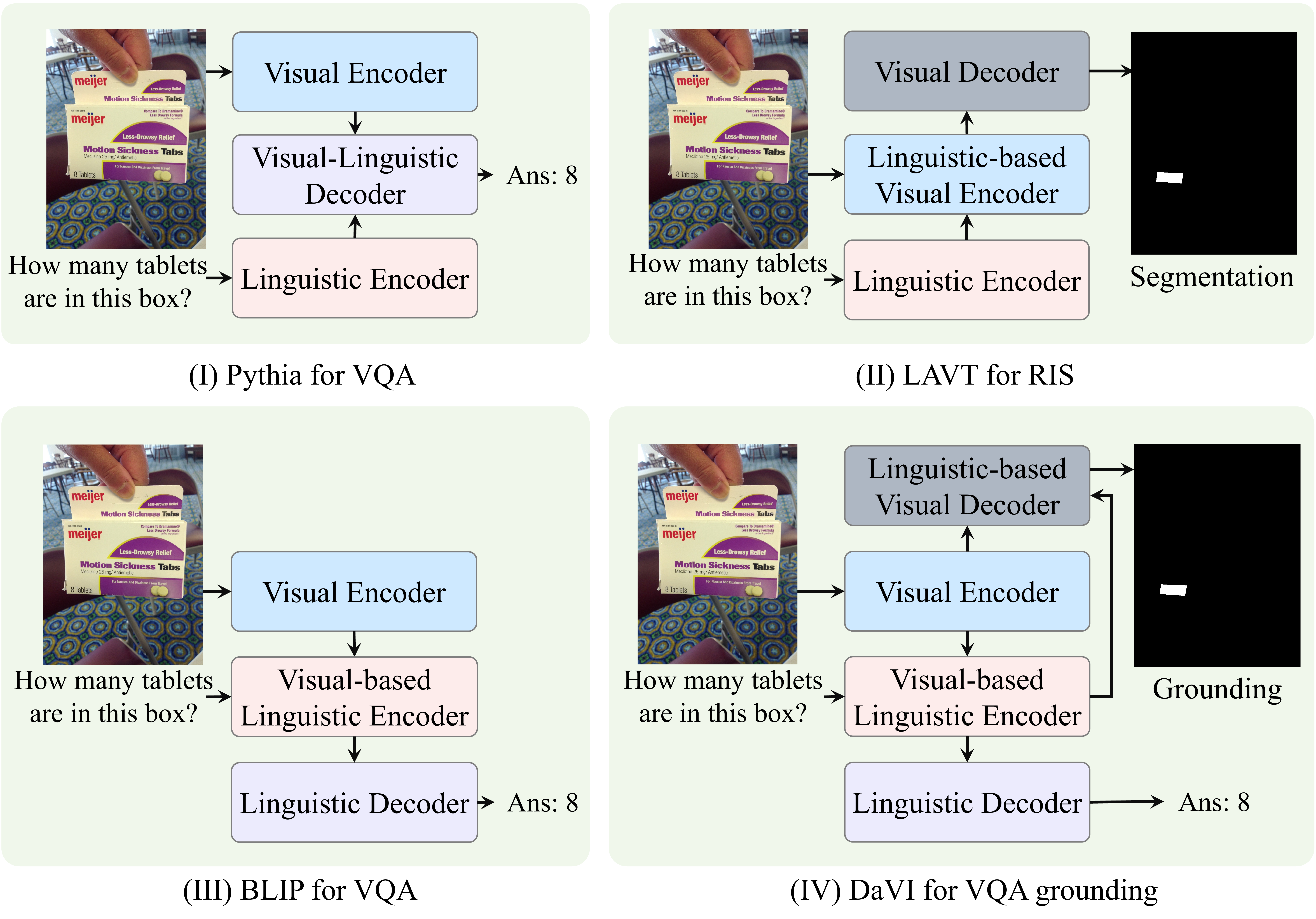}
   \vspace{-2em}
   \caption{
       Overview of visual-linguistic interactions: I) Visual-linguistic decoder~\cite{singh2020mmf} for VQA, II) Linguistic-based visual encoder~\cite{yang2021lavt} for refer image segmentation (RIS), III) Visual-based linguistic encoder (VLE)~\cite{li2022blip} for VQA, and IV) our DaVI with VLE and LVE for answer grounding. It predicts an answer along with a mask while taking as inputs an image and a question.
   }
   \vspace{-1em}
   \label{fig:teaser}
\end{figure}

To overcome the challenges of answer grounding, a fundamental but critical perspective is how to reconcile the visual-linguistic interaction for both visual-oriented and linguistic-oriented predictions.
As shown in Fig.~\ref{fig:teaser}, state-of-the-art works inject features from the other modality into the task-oriented modeling, effectively exploiting rich Transformer layers in the task-oriented encoder. 
However, this visual-linguistic interaction is specific to a single-modal output such as linguistic-oriented answering or visual-oriented segmentation.

Accordingly, this paper proposes DaVI which unifies linguistic-oriented answering and visual-oriented grounding into an end-to-end framework. The two key components of DaVI are Visual-based Linguistic Encoder (VLE) and Linguistic-based Visual Decoder (LVD) which will be detailed in the next section.


\section{Method}

As shown in Fig.~\ref{fig:teaser} (IV), DaVI ingests a question along with the corresponding image as input and outputs an evidence mask while predicting the answer. 
For simplicity, we employ a Vision Transformer (ViT), instead of an object detector, to extract image features.
Then, VLE encodes questions with the image references and produces linguistic-oriented evidence features.
The Linguistic Decoder (LD), with the same structure as the BERT, answers visual questions on the basis of linguistic-oriented evidence.
Different from LD, to predict evidence masks, LVD performs another interaction between the visual features from VE and the linguistic-oriented evidence.



\paragraph{Visual-based Linguistic Encoder (VLE).}
We interleave cross-attention (CA) layers in between the self-attention layers and the feed-forward networks of BERT to form the VLE module.
The CA layers attend to visual references using linguistic features as queries. 
These visual-linguistic interactions in VLE provide multi-modal evidence for answer decoding in LD.
However, due to the language-oriented modeling, these features need to be remapped into visual grids by LVD to decode visual evidence.


\paragraph{Linguistic-based Visual Decoder (LVD).}
LVD can be regarded as an ``identical'' structure of the CA layer in VLE but with different inputs: it uses visual features as queries to attend to linguistic cues from VLE.
Following the best practices in computer vision, we collect multi-level visual features generated by VE and feed them into LVD.
Finally, a simple convolutional segmentor is exploited to predict the visual mask for answer grounding.


\section{Experiment}

\paragraph{Dataset.}
The dataset of VizWiz answer grounding contains 6494, 1131 and 2373 samples in the \textit{train}, \textit{val} and \textit{test} set, respectively. 
Evaluations are conducted by the challenge server on the \textit{test} set.

\paragraph{Implementation Details.}
In our experiments, VE, VLE, and LD  are pretrained on the web-scale dataset~\cite{li2022blip}.
In the training phase, we keep an exponential moving average copy of parameters for inference.
We also ensemble the results from models trained with different hyperparameters.



\begin{table}[!t]
{
\caption{Ablation study on \textit{test} set of VizWiz Grounding track.}
\label{tab:ablation}
\vspace{-.5em}
\small
\begin{tabularx}{\linewidth}{ X | c | c}
\toprule[1pt]
Variant &  IoU (\%) & $\Delta$ (\%) \\
\midrule[.5pt]
LE + LVE + VD (w/o answer) & 64.28 & {--} \\
VE+VLE+LVD+LD & 68.52 & +4.24 \\
\midrule[.5pt]
VE+VLE+LVD+LD (w/ ensemble) & \textbf{70.57} & +2.05 \\
\bottomrule[1pt]
\end{tabularx}}
\vspace{-.5em}

\end{table}

\begin{table}[!t]
{
\centering
\caption{Results on \textit{test} set of VizWiz Grounding track.}
\label{tab:leaderboard}
\vspace{-.5em}
\small
\begin{tabularx}{\linewidth}{ c | X |  c}
\toprule[1pt]
Rank&  Team&  IoU (\%) \\
\midrule[.5pt]
 \textbf{1}  &  \textbf{Aurora (ours)}  &  \textbf{70.57}  \\
2  &hsslab  &70.15  \\
3  &MGTV  &69.70  \\
4  &UD VIMS Lab  &67.55  \\
5  &MindX  &66.92  \\
6  &boostboom  &64.49  \\
7  &Black  &58.44  \\
\bottomrule[1pt]
\end{tabularx}}
\vspace{-.5em}
\end{table}

\paragraph{Results.} 
As shown in Tab~\ref{tab:ablation}, compared to the single interaction paradigm, DaVI successfully achieved a $4.24\%$ IoU improvement, thanks to the design of dual visual-linguistic interactions.
Tab.~\ref{tab:leaderboard} shows public results on VizWiz answer grounding track, where DaVI ranked the 1$^{st}$ place, substantially outperforming other methods.

\section{Conclusion}

We propose DaVI for answer grounding, which utilizes the VLE and LVD for visual-linguistic interaction to unify the linguistic-oriented answering and visual-oriented grounding.
DaVI ranked the 1$^{st}$ place on the track of VizWiz answer grounding challenge.

{\small
\bibliographystyle{ieee_fullname}
\bibliography{egbib}

\begin{thebibliography}{1}\itemsep=-1pt

\bibitem{chen2022grounding}
Chongyan Chen, Samreen Anjum, and Danna Gurari.
\newblock Grounding answers for visual questions asked by visually impaired
  people.
\newblock {\em arXiv preprint arXiv:2202.01993}, 2022.

\bibitem{KhanKDGLS21:found}
Aisha~Urooj Khan, Hilde Kuehne, Kevin Duarte, Chuang Gan, Niels da
  Vitoria~Lobo, and Mubarak Shah.
\newblock Found a reason for me? weakly-supervised grounded visual question
  answering using capsules.
\newblock In {\em CVPR}, pages 8465--8474, 2021.

\bibitem{li2022blip}
Junnan Li, Dongxu Li, Caiming Xiong, and Steven Hoi.
\newblock Blip: Bootstrapping language-image pre-training for unified
  vision-language understanding and generation.
\newblock {\em arXiv preprint arXiv:2201.12086}, 2022.

\bibitem{singh2020mmf}
Amanpreet Singh and et al.
\newblock Mmf: A multimodal framework for vision and language research.
\newblock \url{https://github.com/facebookresearch/mmf}, 2020.

\bibitem{yang2021lavt}
Zhao Yang, Jiaqi Wang, Yansong Tang, and et al.
\newblock Lavt: Language-aware vision transformer for referring image
  segmentation.
\newblock In {\em CVPR}, 2022.

\end{thebibliography}
}

\end{document}